\documentclass[conference]{IEEEtran}
\IEEEoverridecommandlockouts
\usepackage{cite}
\usepackage{amsmath,amssymb,amsfonts}
\usepackage{graphicx}
\usepackage{textcomp}
\usepackage{xcolor}

\usepackage{soul}

\usepackage{subcaption}
 \usepackage{amssymb}
\usepackage{mathtools}
\usepackage{textcomp}
\usepackage{stfloats}
\usepackage{xcolor}
\usepackage{booktabs} 
\usepackage{multirow}
\usepackage{url}
\usepackage{paralist}
\usepackage[noend]{algpseudocode}
\usepackage{balance}
\usepackage[ruled]{algorithm2e}
\usepackage{bbding}
\usepackage{pifont}
\usepackage[inline]{enumitem}
\usepackage{amsthm}

\usepackage[english]{babel}
\newtheorem{theorem}{Theorem}[section]
\newtheorem{definition}{Definition}[section]
\newtheorem{conjecture}{Conjecture}[section]
\newtheorem{corollary}{Corollary}[theorem]

\def\BibTeX{{\rm B\kern-.05em{\sc i\kern-.025em b}\kern-.08em
    T\kern-.1667em\lower.7ex\hbox{E}\kern-.125emX}}
\begin{document}

\title{On the Compressibility of Quantized Large Language Models
}

\author{
  \IEEEauthorblockN{Yu Mao\IEEEauthorrefmark{1}, Weilan Wang\IEEEauthorrefmark{1}, Hongchao Du\IEEEauthorrefmark{1}, Nan Guan\IEEEauthorrefmark{1} and Chun Jason Xue\IEEEauthorrefmark{2}}
  \IEEEauthorblockA{\IEEEauthorrefmark{1}City University of Hong Kong \\ Email: \{yumao7-c, weilawang2-c, hongcdu2-c\}@my.cityu.edu.hk, nanguan@cityu.edu.hk}
  \IEEEauthorblockA{\IEEEauthorrefmark{2}MBZUAI Email: jason.xue@mbzuai.ac.ae}
}

\maketitle

\begin{abstract}
Deploying Large Language Models (LLMs) on edge or mobile devices offers significant benefits, such as enhanced data privacy and real-time processing capabilities. However, it also faces critical challenges due to the substantial memory requirement of LLMs. Quantization is an effective way of reducing the model size while maintaining good performance. However, even after quantization, LLMs may still be too big to fit entirely into the limited memory of edge or mobile devices and have to be partially loaded from the storage to complete the inference. 
In this work, we take a preliminary step of studying applying data compression techniques to reduce data movement and thus speed up the inference of quantized LLM on memory-constrained devices. In particular, we discussed the compressibility of quantized LLMs, the trade-off between the compressibility and performance of quantized LLMs, and opportunities to jointly optimize both of them.
\end{abstract}


\section{Introduction}

%
%
Constrained memory becomes a bottleneck for Large Language Models (LLMs) deployments~\cite{sheng2023flexgen, alizadeh2023llm, dao2022flashattention, zhang2023dissecting, kaddour2023}.
For example, most mobile devices are equipped with 4-12 GB of memory, which hinders running large models effectively.
Some approaches have been proposed to alleviate this burden, including quantization~\cite{xiao2023Smoothquant, frantar2022gptq, liu2023llm, dettmers2022llm}, activation recomputation~\cite{korthikanti2023reducing, jain2020checkmate, smith2022using}, and offloading~\cite{ren2021zero, alizadeh2023llm, sheng2023flexgen}. In this paper, we provide another perspective that has been overlooked in solving this problem. 
%

%

In 2016, DeepCompression~\cite{han2015deep} introduced a quantization method for compressing models and also applied Huffman coding to the quantized weights. However, given the model sizes at that time, entropy coding had a negligible effect. In DeepCompression, Huffman coding achieved a mere 1.2x compression ratio in addition to the 4x ratio from quantization. Consequently, entropy coding has been largely overlooked in later developments of model compression techniques~\cite{xiao2023Smoothquant, frantar2022gptq, liu2023llm, dettmers2022llm}.



Large-language models create opportunities for entropy coding due to increased redundancy in the model. This paper shows that by applying Int8 quantization and a lossless entropy coder to large-language models, it is possible to compress \textbf{model weights by 8x} and \textbf{activations by 16x} without compromising accuracy. Quantization alone provides a 4x compression, while the added entropy coding contributes an additional 2x for weights and 4x for activations.

%

Entropy coding's introduction is challenging due to its reliance on prior quantization methods, which affect both compression ratios and model accuracy.
%
This paper examines how quantization affects a model's compressibility and accuracy, with entropy as the linking factor. 
Assuming the effective knowledge is uniformly distributed across a matrix, selecting a quantization method to enhance compression typically results in reduced accuracy of the quantized model. This trade-off arises as higher compression entails information loss, which is inevitable given the uniform distribution of information.
The hypothesis is tested by comparing two common quantization methods with varying distributions after quantization.

Fortunately, effective information often exhibits locality in the matrix. Recent research on large language models (LLMs)~\cite{dettmers2022llm, lin2023awq, yao2022zeroquant, xiao2023Smoothquant, wei2023outlier} has found that some channels in their matrices, known as outliers, are more crucial for accuracy than others. The key to simultaneously achieving high accuracy and compressibility in quantized models lies in preserving outlier information while reducing non-outlier information. We explore an LLM quantization technique that preserves outlier data while reducing non-outlier data, resulting in similar accuracy and higher compression ratios compared to LLM quantization methods that preserve all data informations.

\begin{figure}[t!]
\centerline{\includegraphics[width=\linewidth]{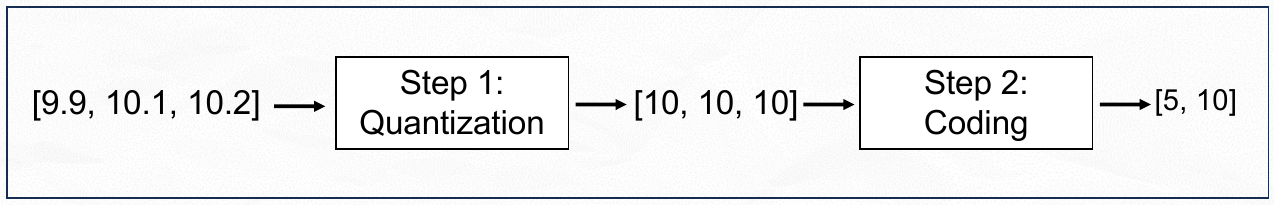}}
\caption{A complete lossy compression process involves two steps: quantization and entropy coding. The latter has long been overlooked in current model compression research. For OPTs, INT8 quantization achieves 4x compression ratio, while entropy coding achieves 2x for weights and 4x for activations.}
\label{fig:two_stage}
\end{figure}

\section{Related work}

\textbf{Quantization.} 
Quantization reduces LLM model size. LLM.int8(~\cite{dettmers2022llm}) uses 8-bit vector-wise quantization and mixed INT8/FP16 precision. 
AWQ~\cite{lin2023awq} finds weights' importance for LLMs' performance. ZeroQuant~\cite{yao2022zeroquant} uses token-wise activation quantization and group-wise weight quantization. Smoothquant~\cite{xiao2023Smoothquant} smooths activation magnitudes for better quantization. Outlier Suppression+~\cite{wei2023outlier} extends understanding of asymmetric distribution in outliers on specific channels.

\textbf{Modern Data Compression Techniques.}
Modern compression techniques like Deflate~\cite{deutsch1996deflate}, Gzip~\cite{b4}, and Zstandard~\cite{b8} combine Huffman~\cite{huffman1952method} and LZ~\cite{ziv1977universal} algorithms with unique designs. FSE~\cite{duda2013asymmetric} is a new kind of entropy encoder based on ANS theory that achieves high-speed compression accuracy. Recently, neural-network-based compressors~\cite{mao2022accelerating, mao2022trace, mao2023faster} have achieved remarkable ratios but unsatisfactory speeds. Therefore this paper only focuses on simple entropy coders.

\section{LLM's compressibility and accuracy}

\begin{figure}[t!]
\centerline{\includegraphics[width=\linewidth]{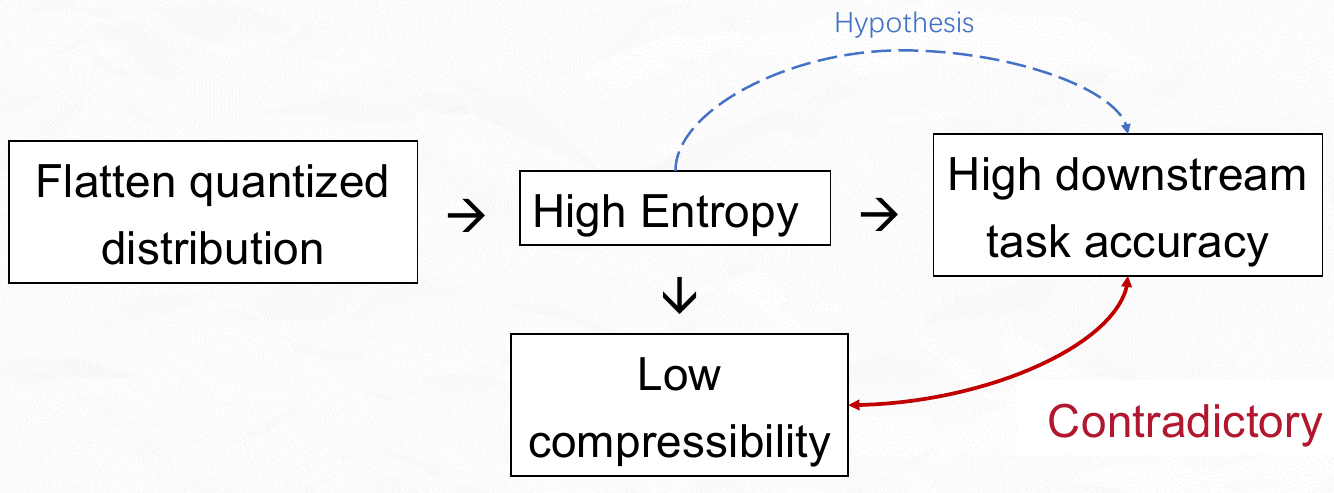}}
\caption{The relationship between quantized weights / activations, entropy, accuracy, and compressibility. Achieving high accuracy and high compressibility is contradictory in theory.}
\label{fig:relationship}
\end{figure}

The section defines information entropy and its relation to data compressibility, then proposes a conjecture based on recent research~\cite{compagi, deletang2023language, yu2024white} that suggests higher accuracy in quantized models leads to lower compressibility when the effective information is evenly distributed in matrix. An experimental comparison of tensor-wise and channel-wise quantization methods is provided to justify this conjecture.




\subsection{Compressibility and Accuracy}

\subsubsection{Compressibility and Entropy}

In information theory, the entropy of a random variable is the average level of "information" inherent to the variable's possible outcomes~\cite{shannon1948mathematical}. 
\begin{definition}
Given a discrete random variable $X$, which takes values in the alphabet $\mathcal{X}$ and is distributed according to $p: \mathcal{X} \rightarrow[0,1]$, the entropy is
\begin{align}
\mathrm{H}(X):=-\sum_{x \in \mathcal{X}} p(x) \log p(x)
\end{align}
where $\Sigma$ denotes the sum of the variable's possible values. 
\end{definition}

\begin{theorem}
The entropy of data increases with the uniformity of its distribution, reaching a maximum when the distribution is completely uniform.~\cite{shannon1948mathematical}.
\end{theorem}

Entropy measures the unpredictability of data; higher entropy indicates greater difficulty in prediction and compression.

\begin{theorem}
Higher entropy leads to lower compressibility and vice versa~\cite{shannon1948mathematical}.
\end{theorem}

\subsubsection{The relationship between quantized model accuracy and entropy}
\label{sec:acc_com}

This section explores the connection between quantized model accuracy and entropy. 
%
%
We need to prioritize the assumption that information is uniformly distributed across the matrix, representing an ideal scenario. This means every piece of data within the matrix holds equal importance and equally influences model accuracy. Our goal is to ensure all data information is fully preserved. Consequently, a higher entropy in the quantized matrix signifies more knowledge, which translates into greater accuracy for the quantized model. However, by definition, increased entropy also results in reduced compressibility.

\begin{conjecture}
With the given assumption, higher matrix entropy leads to improved model accuracy after quantization.
\end{conjecture}

Based on the above conjecture, we make a claim about the relation between accuracy and compressibility. 

\begin{corollary}
    With the given assumption, the higher compressibility of the quantized model indicates lower quantized model accuracy.
\end{corollary}

We re-emphasize that the model compression method outlined in this paper involves two steps: \textbf{quantization and entropy coding}. 
%
The impact of different quantization methods on entropy coding performance will be discussed in the upcoming sections. Note that all compression ratios reported in subsequent sections are in addition to a 4x reduction from Int8 quantization, which means the total compression ratio is $4 \times C$, where $C$ is the additional compression ratio achieved by entropy coding.





\subsection{Experimental Justification}

The comparison of tenor-wise and channel-wise quantization is used as an instance to justify the above corollary with experimental results. 
Tensor-wise and channel-wise quantization are two commonly used model quantization methods. Tensor-wise quantization selects max/min values for the entire matrix range, resulting in uneven quantized distribution. Channel-wise quantization selects max/min values for each row/column, resulting in a more flatten quantized distribution.
%



\begin{figure}[t!]
\centerline{\includegraphics[width=0.66\linewidth]{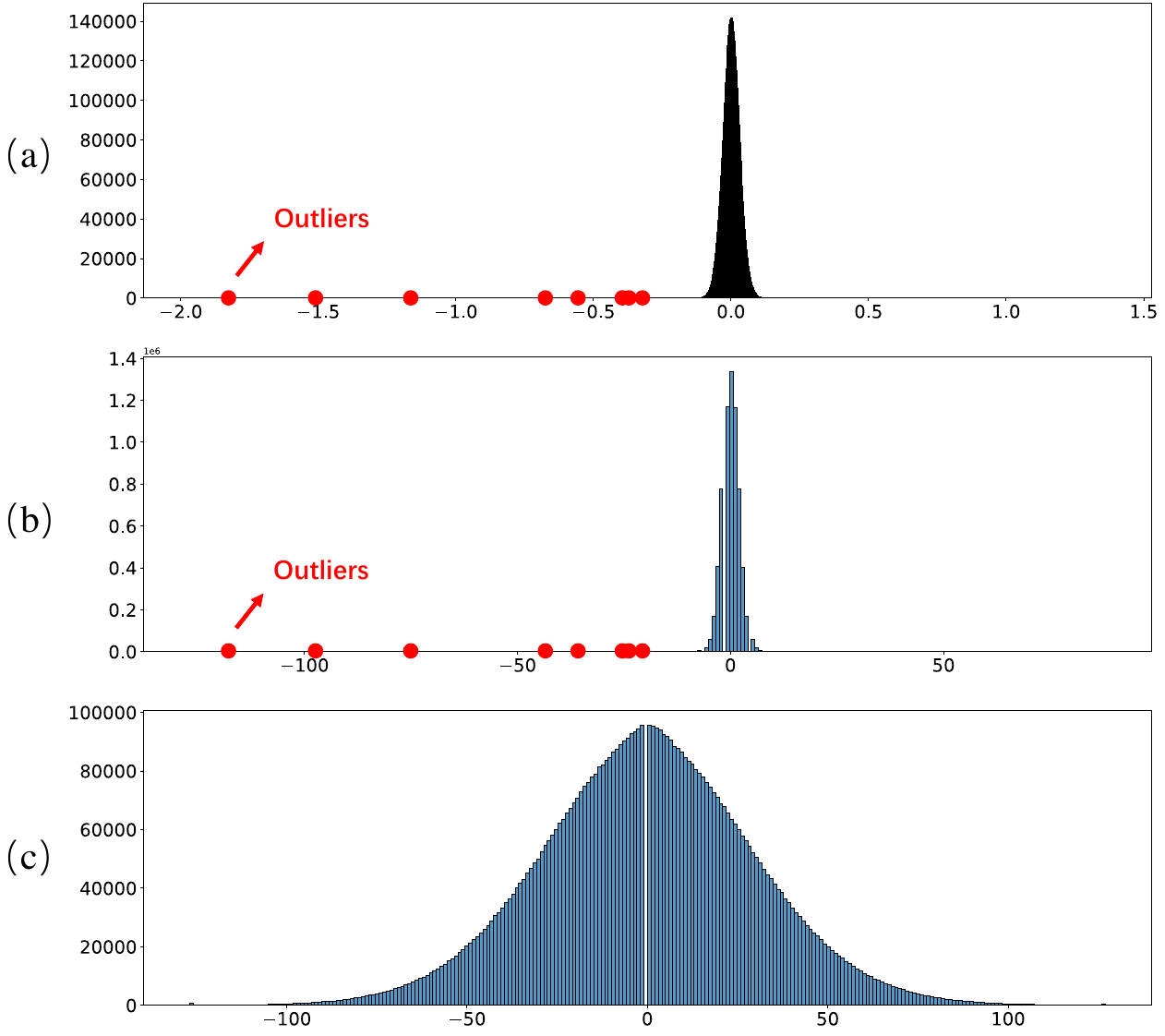}}
\caption{Weight distributions of the original data matrix, tensor-wise and channel-wise quantized matrix.
}
\label{fig:distribution}
\end{figure}

Channel-wise quantized matrices have more information but are less compressible, while tensor-wise quantized matrices have less information but are more compressible due to different distribution characteristics. Fig.~\ref{fig:distribution} shows the distribution of $mlp.c\_proj.weight$ in the GPT2-large model. 
Fig.~\ref{fig:distribution}(a) shows the non-uniform distribution of original floating-point numbers, including outliers. Fig.~\ref{fig:distribution}(b) depicts the leptokurtic distribution of weights after tensor-wise quantization, with most values confined to -10 to 10. In contrast, channel-wise quantization results in a platykurtic distribution shown in Fig.~\ref{fig:distribution}(c). This contributes to higher weight entropy and reduced compressibility (Section 4.1).


Naive tenor-wise and channel-wise quantization methods are implemented on GPT2-large model to show the conclusion. 
The overall compression ratio is shown in Tab.~\ref{tab:gpt2}. We can conclude that the channel-wise quantization indeed has lower compressibility than tensor-wise quantization with higher accuracy, therefore demonstrating our conclusion.




\begin{figure}[t!]
\centering
\begin{subfigure}{0.44\textwidth}
    \includegraphics[width=\textwidth]{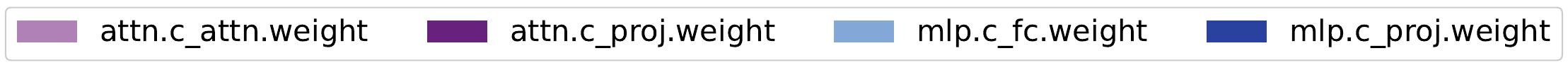}
\end{subfigure}
\begin{subfigure}{0.23\textwidth}
    \includegraphics[width=\textwidth]{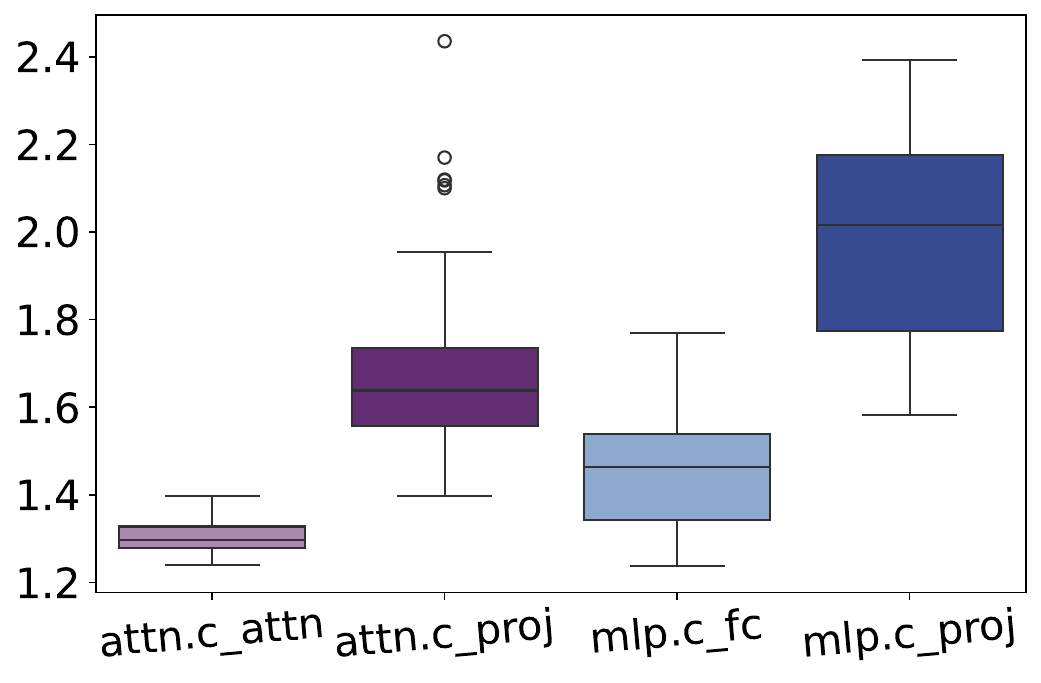}
    \caption{Tensor-wise compression ratio statistic features.}
    \label{fig:t_1}
\end{subfigure}
\begin{subfigure}{0.23\textwidth}
    \includegraphics[width=\textwidth]{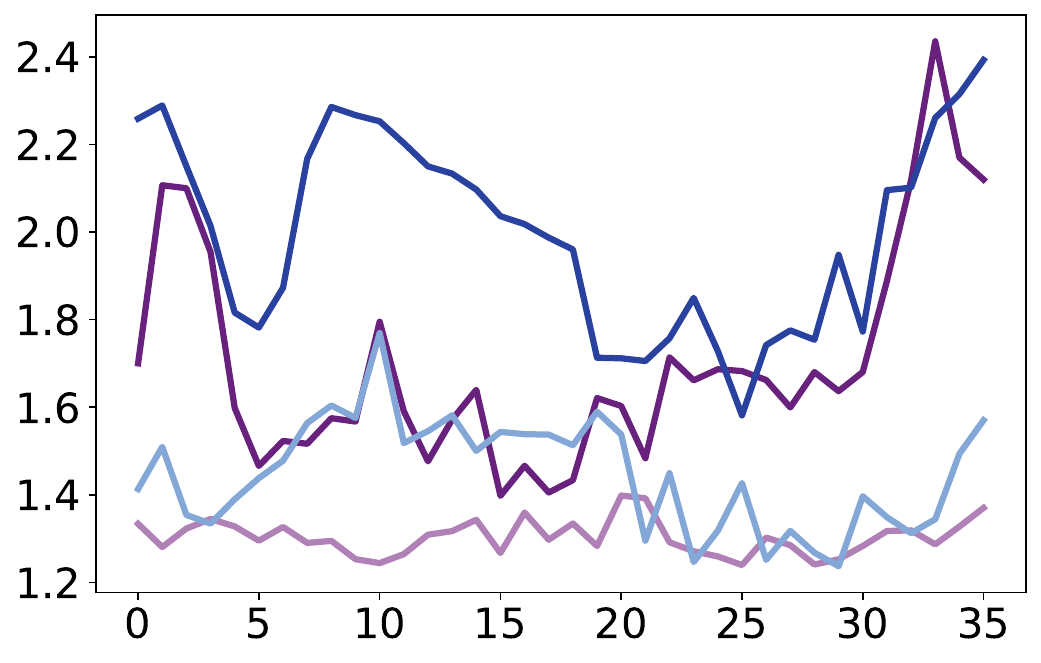}
    \caption{Tensor-wise each layer compression ratios.}
    \label{fig:t_2}
\end{subfigure}
\begin{subfigure}{0.23\textwidth}
    \includegraphics[width=\textwidth]{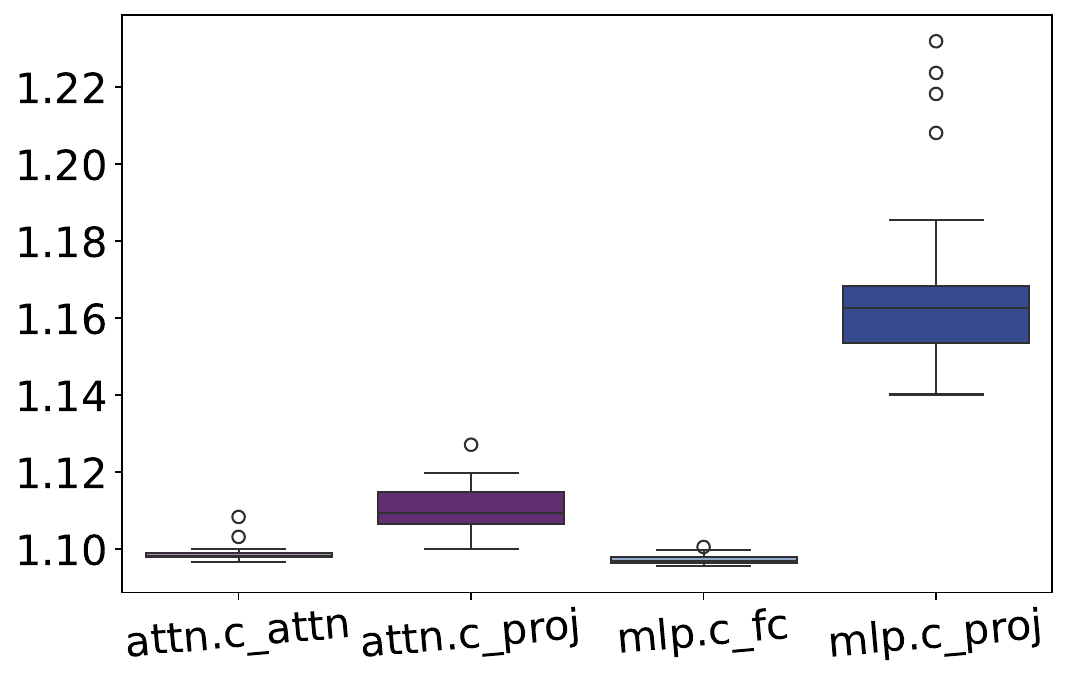}
    \caption{Channel-wise compression ratio statistic features.}
    \label{fig:c_1}
\end{subfigure}
\begin{subfigure}{0.23\textwidth}
    \includegraphics[width=\textwidth]{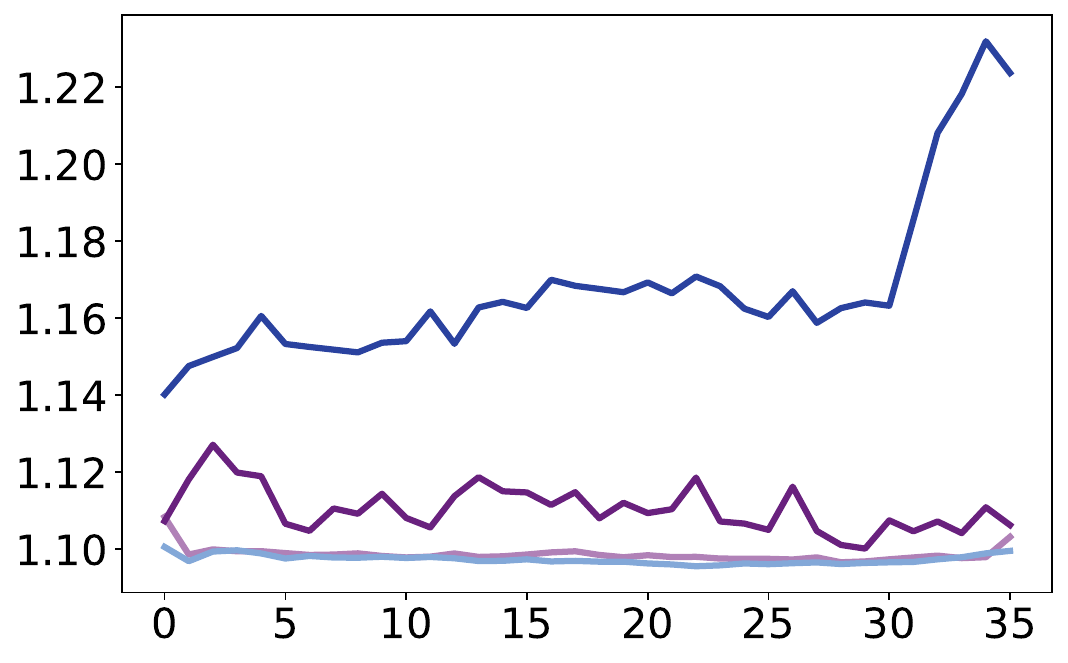}
    \caption{Channel-wise each layer compression ratios.}
    \label{fig:c_2}
\end{subfigure}
\caption{Compression ratios on tensor-wise and channel-wise quantized model.}
\label{fig:per_tensor}
\end{figure}

\begin{table}[htbp]
\caption{Layer Average Entropy Coding Compression Ratio in GPT2-large.}
\begin{center}
\normalsize
\setlength\tabcolsep{4pt}
\begin{tabular}{|c|c|c|c|c|}
\hline
\textbf{Avg. CR.} & attn.c\_attn  & attn.c\_proj &  mlp.c\_fc & mlp.c\_proj\\
\hline
Tensor-wise & 1.09 & 1.11 & 1.09 & 1.17 \\
\hline
Channel-wise & 1.31 & 1.71 & 1.45 & 1.99 \\
\hline
\end{tabular}
\label{tab:gpt2}
\end{center}
\end{table}

We further give out the detailed compression performance statistics of the two quantization methods.
%
From Fig.~\ref{fig:per_tensor}, we can draw the following conclusions:
\begin{itemize}
    \item Under per-tensor quantization, model weights can still be compressed to 1.2x to 2.4x after INT8 quantization. 
    \item Different layers have varying compression ratios, such as $attn.c\_attn.weight$ averaging at 1.3 and $mlp.c\_proj.weight$ being at least 1.6 or higher. 
    \item The first and last layers generally have higher compression ratios compared to the middle layer. 
    \item Channel-wise quantization reduces following compressibility significantly, resulting in a compression rate of only 1.1 to 1.2 times. 
\end{itemize}


\section{Opportunities for attaining both metric excellence}
The statement in Sec.~\ref{sec:acc_com} suggests a challenging outcome: as accuracy improves, compressibility diminishes. 
We will discuss the opportunities to attain good performance on both accuracy and compressibility.
Specifically, by employing SmoothQuant, the cutting-edge LLM quantization technique, we show that with careful design, it's possible to simultaneously attain impressive results in both areas.



\subsection{Outlier-aware LLM Quantization Method}


%
%
The opportunity to achieve good performance on both compressibility and quantized LLM accuracy lies in leveraging the effective information locality. The assumption in Sec.~\ref{sec:acc_com} that information is uniformly distributed across the matrix may not hold, as effective information often shows local patterns in current Large Language Models (LLMs). An example of this is the presence of outliers. Recent studies have shown that not all channels in large language models (LLMs) are equally important~\cite{xiao2023Smoothquant, frantar2022gptq, liu2023llm, dettmers2022llm}. Specifically, only a select few channels within each matrix—referred to as outliers—have a significant impact on the model's performance. In contrast, the remaining channels, deemed non-outliers, contribute minimally.
We suggest that by selectively quantizing data, prioritizing important information and compressing less critical data, we can simultaneously attain high compression rates and maintain accuracy.

\begin{figure}[t!]
\centerline{\includegraphics[width=0.8\linewidth]{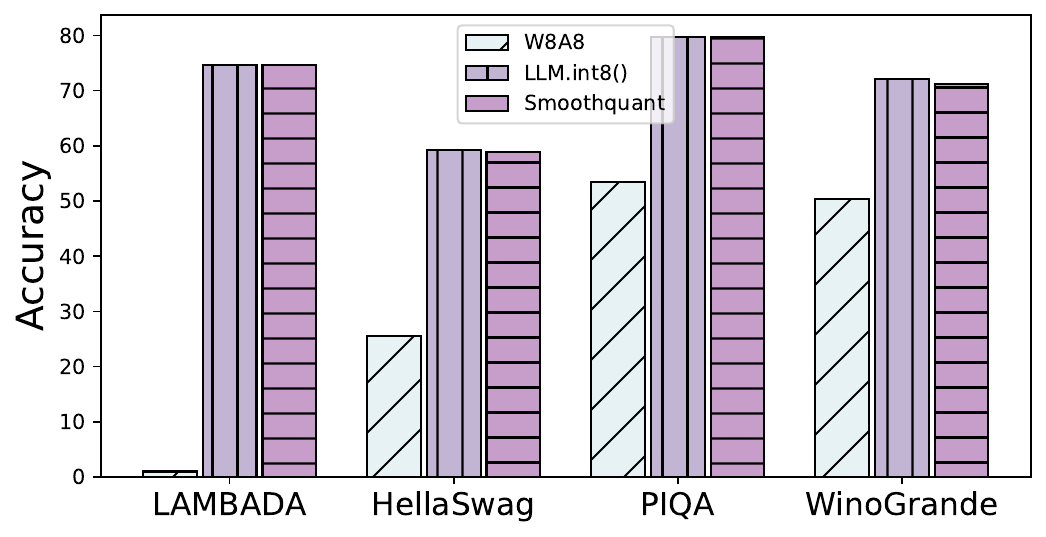}}
\caption{Accuracy on four downstream tasks of W8A8 (Tensor-wise), Smoothquant (Tensor-wise), and LLM.int8()  (Channel-wise).}
\label{fig:acc}
\end{figure}

Smoothquant, an outlier-aware tensor-wise quantization method for LLMs, matches the accuracy of channel-wise quantization on downstream tasks (Fig.~\ref{fig:acc}) and maintains a high entropy coding compression ratio (Tab~\ref{tab:aw_comp}). Smoothquant operates by dividing input activation with a per-channel smoothing factor, and to keep the calculations equivalent, it multiplies the weights by the inverse of this smoothing factor.

%

\begin{equation}
\mathrm{Y}=\left(\mathrm{X} \operatorname{diag}(\mathrm{s})^{-1}\right) \cdot(\operatorname{diag}(\mathrm{s}) \mathrm{W})=\hat{\mathrm{X}} \hat{\mathrm{W}}
\end{equation}

Such smoothing factor is calculated from values of weights and activations. 


\begin{equation}
\mathrm{s}_j=\frac{\max \left(\left|\mathrm{X}_j\right|\right)^\alpha}{\max \left(\left|\mathrm{W}_j\right|\right)^{1-\alpha}}
\end{equation}

This step transfers a portion of the activation numeric values to the weights, ensuring that outliers maintain a larger value in both the activation and weight matrices. This ensures that during quantization, these outliers and their corresponding weights preserve maximum information by receiving more effective bits due to their larger original values. 
For example, when an outlier activation is 100 and its weight is 1, the operation adjusts both values to 10. Given that 10 is still high relative to typical activations (the original outlier is over 100 times larger than non-outliers) and weights (usually between 0 and 1), post-Int8 quantization will push them closer to the maximum value of 127, resulting in more effective bits.

\begin{table}[t]
\caption{Layer Average Entropy Coding Compression Ratio in OPTs. 
}
\begin{center}
\normalsize
\setlength\tabcolsep{4pt}
\begin{tabular}{|c|c|c|c|}
\hline
\textbf{Avg. CR.} & OPT-1.3b  & OPT-6.7b &  OPT-13b\\
\hline
Tensor-wise Activation  & 4.32 & 4.35 & 4.98 \\
\hline
Smoothquant Activation  & 3.14 & 3.23 & 3.16\\
\hline
LLM.int8() Activation  & 1.84 & 1.87 & 1.87\\
\hline
Tensor-wise Weight & 1.54 & 1.36 & 1.32\\
\hline
Smoothquant Weight & 1.64 & 1.52 & 1.46 \\
\hline
LLM.int8() Weight & 1.16 & 1.14 & 1.13 \\
\hline
\end{tabular}
\label{tab:aw_comp}
\end{center}
\end{table}

Tab.~\ref{tab:aw_comp} shows the average compression ratio for LLM.int8(), Smoothquant, and tensor-wise quantization on OPT models. For the activation processed by Smoothquant, the amount of information increased significantly (the compression rate increased by 25\%$\sim$30\%), while the amount of weight information only increased by $\sim$7\%. Another interesting phenomenon is that compared to the compression rate of weight which is only about 1.5x, the compressibility of activation is obviously better and can reach 5x.



\section{Experiments}

To further extend our conclusion to practical LLM application scenarios, this section explore the compression effect on a widely utilized LLM series, OPT models~\cite{zhang2022opt} under two state-of-the-art LLM quantization methods, Smoothquant (INT8, \textit{tensor-wise})~\cite{xiao2023Smoothquant} and LLM.int8() (INT8, \textit{channel-wise})~\cite{dettmers2022llm}.

\subsection{Compression Ratio Under Different Entropy Coders} 

Huffman-Coding~\cite{huffman1952method}, FSE Coding~\cite{duda2013asymmetric}, and Zstandard~\cite{b8} are selected as compression baselines. As illustrated, under the state-of-the-art compression method, the model after tensor-wise Smoothquant has a compression rate of more than 1.5 times, while the channel-wise LLM.int8() quantized weight only has a compression rate of 1.1$\sim$1.3 times. 

\begin{table}[htbp]
\caption{Compression Ratio for Overall Quantized Model under Different Entropy Coders.}
\begin{center}
\normalsize
\setlength\tabcolsep{4pt}
\begin{tabular}{|c|c|c|c|c|}
\hline
\textbf{SmoothQuant} & OPT-1.3b &  OPT-6.7b & OPT-13b & OPT-30b\\
\hline
Zstd & 2.22 & 1.78 & 1.66 & 1.69\\
\hline
FSE & 2.11 & 1.77 & 1.66 & 1.68\\
\hline
Huffman & 2.15 & 1.76 & 1.66 & 1.69\\
\hline
\textbf{LLM.int8()} & OPT-1.3b &  OPT-6.7b & OPT-13b & OPT-30b\\
\hline
Zstd & 1.3 & 1.2 & 1.18 & 1.11\\
\hline
FSE & 1.31 & 1.2 & 1.18 & 1.11\\
\hline
Huffman & 1.3 & 1.2 & 1.18 & 1.11\\
\hline
\end{tabular}
\label{tab:smooth}
\end{center}
\end{table}

\subsection{Efficiency Report}

\textbf{Compression Speed.} 
Three compression methods were compared in terms of their speeds. FSE coding had the slowest speed, especially for decompression. Huffman Coding had a compression speed twice as fast as FSE coding and a decompression speed of 1.3G per second. Zstd achieved the best overall performance with the fastest compression and decompression speeds, while maintaining a satisfactory compression rate. Its decompression speed was twice that of Huffman coding at 2.3GB per second (Tab.~\ref{tab:speed}).


\begin{table}[htbp]
\caption{Entropy Coding and Decoding Speed.}
\begin{center}
\normalsize
\setlength\tabcolsep{4pt}
\begin{tabular}{|c|c|c|c|}
\hline
\textbf{Speed(MB/s)} & zstd &  FSE & Huffman\\
\hline
Comp. Speed & 710 & 325 & 600\\
\hline
Decomp. Speed & 2300 & 440 & 1350\\
\hline
\end{tabular}
\label{tab:speed}
\end{center}
\end{table}



\textbf{Model Loading Time Reduction in Real-World Scenario.} To show the effect of compression on model loading, we measure loading time before and after using compression. We deploy the experiments on a testbed with an 8-core Intel Core i7-9700 CPU (3.00Ghz) and 64GB DDR4 DRAM. A 512GB KIOXIA XG6 SSD is utilized for storage with performance up to 3180 MB/s and 2800 MB/s for sequential read and write, respectively.
%
%
%
Table~\ref{tab:loading} shows the results. Mmap performs at half the level of read system call due to 4KB parameter loading and limited storage bandwidth utilization. Compression reduces loading time by 41-55\% with mmap and 40-60\% with read, similar to using multiple threads for loading time reduction.


\begin{table}[htbp]
\caption{Loading Time Before and After Entropy Coding.}
\begin{center}
\normalsize
\setlength\tabcolsep{4pt}
\begin{tabular}{|c|c|c|c|c|}
\hline
\textbf{Time(s)} & OPT-1.3b  & OPT-6.7b &  OPT-13b & OPT-30b\\
\hline
mmap-before & 1.02 & 4.52 & 8.45 & 19.17\\
\hline
mmap-after & 0.46 & 2.51 & 5.0 & 11.32\\
\hline
read-before & 0.55 & 2.46 & 4.6 & 10.52\\
\hline
read-after & 0.25 & 1.39 & 2.76 & 6.28\\
\hline
\end{tabular}
\label{tab:loading}
\end{center}
\end{table}

\section{Conclusion}

The paper explores the use of compression to reduce data volume in large language model (LLM) calculations, finding that quantized LLMs are highly compressible. The level of quantization granularity influences this compressibility by altering weight distribution, which is analyzed using information theory. Experiments with two advanced LLM quantization methods confirm these results. Furthermore, a loading experiment shows that compression can double the data transfer rate between flash memory and DRAM. 


\bibliographystyle{ieeetr}
\bibliography{main}

\end{document}